\documentclass[acmtog,nonacm]{acmart}

\AtBeginDocument{%
  \providecommand\BibTeX{{%
    \normalfont B\kern-0.5em{\scshape i\kern-0.25em b}\kern-0.8em\TeX}}}

\copyrightyear{2024}
\acmYear{2024}
\setcopyright{rightsretained}
\acmConference[SA Conference Papers '24]{SIGGRAPH Asia 2024 Conference Papers}{December 3--6, 2024}{Tokyo, Japan}
\acmBooktitle{SIGGRAPH Asia 2024 Conference Papers (SA Conference Papers '24), December 3--6, 2024, Tokyo, Japan}\acmDOI{10.1145/3680528.3687634}
\acmISBN{979-8-4007-1131-2/24/12}

\acmSubmissionID{635}

\usepackage{multirow}
\usepackage{enumitem}
\usepackage{amsmath}
\usepackage{amsfonts}
\usepackage{nicefrac}
\usepackage{mathtools}
\usepackage{wrapfig}
\usepackage{tikz}
\usepackage{xcolor}
\usepackage{colortbl}
\usepackage{booktabs} %
\usepackage{tabularx}
\usepackage{xspace}
\usepackage{bm}
\usepackage{url}
\newcolumntype{Y}{>{\centering\arraybackslash}X}
\newcolumntype{C}[1]{>{\centering\arraybackslash}p{#1}}

\usepackage{custom_style} %


\newif\ifcomments
\commentstrue

\renewcommand{\eqref}[1]{Equation~\ref{eq:#1}}
\newcommand{\figref}[1]{Figure~\ref{fig:#1}}
\newcommand{\secref}[1]{Section~\ref{sec:#1}}

\newcommand{\ours}{\text{SpaceMesh}}

\newcommand{\RR}{\mathbb{R}}

\newcommand{\Mesh}{\mathcal{M}}
\newcommand{\Vertices}{\mathcal{V}}
\newcommand{\Vertex}{v}
\newcommand{\VertexDegree}{D}
\newcommand{\Edges}{\mathcal{E}}
\newcommand{\Edge}{e}

\newcommand{\Halfedge}{h}
\newcommand{\Faces}{\mathcal{F}}
\newcommand{\Next}{\mathcal{N}}
\newcommand{\Graph}{\mathcal{G}}

\newcommand{\VertexPosition}{p}

\newcommand{\hetwin}{\texttt{twin}}
\newcommand{\henext}{\texttt{next}}

\newcommand{\AdjacencyFeature}{x}
\newcommand{\AdjacencyFeatureSpacelike}{x^\textrm{s}}
\newcommand{\AdjacencyFeatureTimelike}{x^\textrm{t}}
\newcommand{\AdjacencyFeatureDim}{k}
\newcommand{\AdjacencyFeatureSpacelikeDim}{{k^\textrm{s}}}
\newcommand{\AdjacencyFeatureTimelikeDim}{{k^\textrm{t}}}
\newcommand{\AdjacencyThreshold}{\tau}
\newcommand{\Distance}{\dsf}
\newcommand{\DistanceEuclidean}{\Distance^\textrm{eu}}
\newcommand{\DistanceSpacetime}{\Distance^\textrm{st}}

\newcommand{\PermutationFeatureRoot}{y^{\textrm{root}}}
\newcommand{\PermutationFeaturePrev}{y^{\textrm{prev}}}
\newcommand{\PermutationFeatureNext}{y^{\textrm{next}}}
\newcommand{\PermutationFeatureDim}{{k^{\textrm{p}}}}

\newcommand{\PermutationMatrix}{\Phi}
\newcommand{\NormalizedPermutationMatrix}{\bar{\Phi}}
\newcommand{\PermutationReduction}{F}

\newcommand{\xvec}{x}

\newcommand{\dsf}{\mathsf{d}}
\newcommand{\eg}{\emph{e.g.}} %
\newcommand{\ie}{\emph{i.e.}} %
\newcommand{\etc}{\emph{etc.}} %
\newcommand{\etal}{\emph{et al.}} %

\begin{document}

\title[SpaceMesh: A Continuous Representation for Learning Manifold Surface Meshes]{SpaceMesh: A Continuous Representation for\\Learning Manifold Surface Meshes}

\author{Tianchang Shen}
\orcid{0000-0002-7133-2761}
\affiliation{%
 \institution{NVIDIA, University of Toronto, Vector Institute}
  \country{Canada}
 }
\email{frshen@nvidia.com}

\author{Zhaoshuo Li}
\orcid{0000-0001-5874-4713}
\affiliation{%
 \institution{NVIDIA}
  \country{USA}
 }
 \email{maxzhaoshuol@nvidia.com}

\author{Marc Law}
\orcid{0000-0001-7767-2313}
\affiliation{%
 \institution{NVIDIA}
  \country{Canada}
}
\email{marcl@nvidia.com}

\author{Matan Atzmon}
\orcid{0009-0001-6998-1033}
\affiliation{%
 \institution{NVIDIA}
  \country{Canada}
}
\email{matzmon@nvidia.com}

\author{Sanja Fidler}
\orcid{0000-0003-1040-3260}
\affiliation{%
 \institution{NVIDIA, University of Toronto, Vector Institute}
  \country{Canada}
}
\email{sfidler@nvidia.com}

\author{James Lucas}
\orcid{0009-0005-4580-7937}
\affiliation{%
\institution{NVIDIA}
\country{UK}
}
\email{jalucas@nvidia.com}

\author{Jun Gao}
\orcid{0000-0002-3521-0417}
\affiliation{%
  \institution{NVIDIA, University of Toronto, Vector Institute}
  \country{Canada}
}
\email{jung@nvidia.com}

\author{Nicholas Sharp}
\orcid{0000-0002-2130-3735}
\affiliation{%
  \institution{NVIDIA}
  \country{USA}
}
\email{nsharp@nvidia.com}
\renewcommand{\shortauthors}{Shen \etal{}}



\begin{abstract}
    Meshes are ubiquitous in visual computing and simulation, yet most existing machine learning techniques represent meshes only indirectly, \eg{} as the level set of a scalar field or deformation of a template, or as a disordered triangle soup lacking local structure.
    This work presents a scheme to directly generate manifold, polygonal meshes of complex connectivity as the output of a neural network.
    Our key innovation is to define a continuous latent connectivity space at each mesh vertex, which implies the discrete mesh.
    In particular, our vertex embeddings generate cyclic neighbor relationships in a halfedge mesh representation, which gives a guarantee of edge-manifoldness and the ability to represent general polygonal meshes.
    This representation is well-suited to machine learning and stochastic optimization, without restriction on connectivity or topology.
    We first explore the basic properties of this representation, then use it to fit distributions of meshes from large datasets.
    The resulting models generate diverse meshes with tessellation structure learned from the dataset population, with concise details and high-quality mesh elements. In applications, this approach not only yields high-quality outputs from generative models, but also enables directly learning challenging geometry processing tasks such as mesh repair.
\end{abstract}

\begin{teaserfigure}
\centering
\vspace{2em}
\includegraphics[trim=0 0 0 0, clip, width=0.97\textwidth]{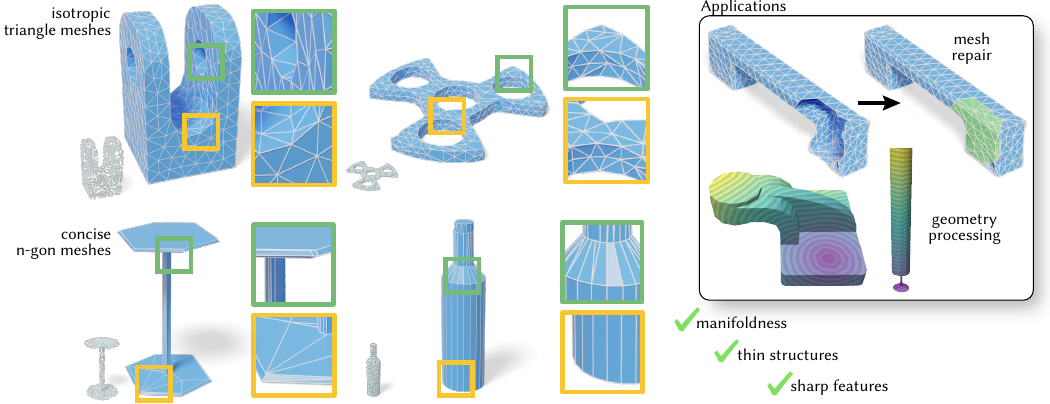}
\caption{
We present a mesh representation that enables learning to directly generate polygonal meshes as the output of a neural network.
\emph{Left:} 3D meshes produced from a generative model trained on a dataset with desirable mesh connectivity.
\emph{Right:} our model can be applied to challenging tasks such as mesh repair, and produces manifold meshes suitable for downstream processing like computing geodesic distance.
\label{fig:teaser}
\vspace{2em}
}
\end{teaserfigure}

\maketitle

\vfill

\pagebreak
\section{Introduction}
\label{sec:intro}

Polygonal meshes play an essential role in computer graphics, favored for their simplicity, flexibility, and efficiency. They can represent surfaces of arbitrary topology with non-uniform polygons, and support a wide range of downstream processing and simulation. Additionally, meshes are ideal for rasterization and texture mapping, making them efficient for rendering. However, the benefits of meshes rely heavily on their quality. For example, meshes with non-manifold connectivity or too many elements may break operations that leverage local structure, or make processing prohibitively expensive.
Consequently, developing automatic algorithms and tools for generating high-quality meshes is an ongoing research focus.

It is no surprise that recent advancements in deep learning have led to growing interest in learning-based mesh creation. 
Generating meshes as output, however, is a notoriously challenging task for machine learning algorithms,
as meshes have a complex combination of continuous and discrete structure.
Not only do mesh vertices and edges form a graph, but mesh faces add additional interconnected structure, and furthermore those faces ought to be arranged locally for manifold connectivity.
Existing approaches range from implicit function isosurfacing~\cite{occnet,gao2022get3d,shen2021dmtet,shen2023flexicubes}, which offers easy optimization and a guarantee of validity at the expense of restricting to a limited family of meshes, to directly generating faces as an array of vertex triplets~\cite{nash2020polygen,alliegro2023polydiff,siddiqui2023meshgpt}, a discrete-first perspective which cannot be certain to respect the constraints of local structure.
This work seeks a solution that offers the best of all worlds: the ease and utility that comes from working in a continuous parameterization, a guarantee to produce meshes with manifold structure by construction, and the generality to represent the full range of possible meshes.

We present \ours, a representation for meshes built on continuous embeddings well-suited for learning and optimization, which guarantees manifold output and supports complex polygonal connectivity. Our approach derives from the halfedge data structure~\citep{weiler1986topological}, which inherently represents manifold, oriented polygonal meshes---the heart of our contribution is a continuous parameterization for halfedge mesh connectivity.

The main idea is to represent mesh connectivity by first constructing a set of edges and halfedges, and then constructing the so-called \emph{next} relationship among those halfedges to implicitly define the faces of the mesh. We introduce a parameterization of edge adjacency and \emph{next} relationships with low-dimensional, per-vertex embeddings. These embeddings, by construction, always produce a manifold halfedge mesh without additional constraints. Moreover, the per-vertex embedding is straightforward to predict as a neural network output and demonstrates fast convergence during optimization. The continuous property of our representation facilitates new architectures for mesh generation, and enables applications like mesh repair with learning.

We validate our representation against alternatives for representing graph adjacency and meshes, and demonstrate superior significantly faster convergence, which is fundamentally important for learning tasks. Combined with a generative model for vertices, we showcase our representation in learning different surface discretization for meshing. Additionally, our representation enables mesh repair via deep learning, simultaneously predicting both vertices and topology.

\section{Related Work}
\label{sec:related}

Initial deep learning-based mesh generation techniques focused on vertex prediction while maintaining fixed connectivity, which are challenging to adapt for complex 3D objects~\cite{Wang2018Pix2Mesh, Groueix2018AtlasNet, Litany2018DeformNet, Ranjan2018Generating3D, Wang2020MeshGraphormer, Tan2020VariationalAutoencoders, Hanocka2020P2Mesh, Chen2019Learning, Zhang2021DatasetGAN, Zhang2020Image}. Although local topology modifications are possible through subdivision~\cite{Wang2018Pix2Mesh,Liu:Subdivision:2020} or remeshing~\cite{palfinger2022continuous}, these methods still struggle to represent general, complex 3D objects. 
Recent methods utilize intermediary representations that are converted into meshes using techniques like Poisson reconstruction on point clouds~\cite{kazhdan2006poisson,Peng2021SAP} or isosurfacing on implicit fields~\cite{shen2023flexicubes,shen2021dmtet,gao2022get3d,magic3d,chen2022ndc}. However, these conversion processes lack precise control over mesh connectivity.

\subsection{Generating Meshes}

Much recent work has specifically studied approaches for generating surfaces meshes in learning-based pipelines.

\paragraph{Volumetric 3D Reconstruction} 
See Point2Surf~\cite{erler2020points2surf}, POCO~\cite{boulch2022poco}, NKSR~\cite{huang2023neural} and BSPNet~\cite{chen2020bspnet}, \etc{} 
These methods focus on reconstructing the geometric shape, rather than the mesh structure; output connectivity is always a marching-cubes mesh (or a union-of-planes in BSPNet).
Our approach instead focuses on fitting particular discrete mesh connectivity structures from data.
\figref{abc_histogram} and \ref{fig:abc_qual} include a few representative methods from this family, although they generally target significantly different goals.
A parallel class of methods leverages Voronoi/Delaunay-based formulations ~\cite{maruani2023voromesh,maruani2024ponq}), but again these focus on fitting a surface's geometric shape, rather than the particular mesh connectivity.

\begin{figure}[h!]
    \centering
    \vspace{2em}
    \includegraphics[width=0.8\columnwidth]{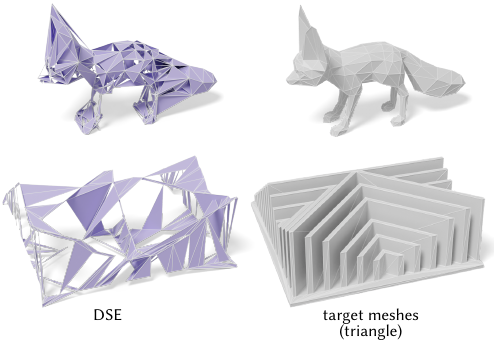}
    \caption{
      Fitting the ground truth connectivity of a single mesh with DSE~\cite{rakotosaona2021learning}. The experiment setting is described in Section~\ref{sec:mesh_to_ours}.
      Here DSE is overfit to encode a single shape, but even then its representation struggles when vertices are sparse and geometry is highly nonconvex.
    }
    \label{fig:dse}
\end{figure}

\paragraph{Direct Mesh Learning} 
See IER~\cite{liu2020meshing}, PointTriNet~\cite{sharp2020pointtrinet}, Delaunay Surface Elements (DSE)~\cite{rakotosaona2021learning},
DMesh~\cite{son2024dmesh}.
Like ours, these approaches aim to directly learn structured mesh connectivity.
However, our approach offers a guarantee of manifoldness, and can encode general polygonal meshes.
Additionally, we demonstrate the ability to encode concise artist/CAD-like tessellation via coupled learning of vertex positions and connectivity, rather than generating faces among a rough uniformly-sampled point set.
Conversely, some of these methods scale to high resolution outputs, compared to our small-medium meshes.
\figref{mesh_to_rep} includes comparisons to DMesh~\cite{son2024dmesh} as a representative method from this family, see also additional results from DSE in \figref{dse} in the same setting.

\paragraph{Sequence Modeling} 
See Polygen~\cite{nash2020polygen}, PolyDiff~\cite{alliegro2023polydiff}, MeshGPT~\cite{siddiqui2023meshgpt}, and the concurrent MeshAnything~\cite{chen2024meshanything}.
These approaches use large-scale architectures to emit a mesh one face or vertex at a time.
Unlike our method, they generally do not offer any guarantees of connectivity or local structure, and all but Polygen produce triangle soup, connecting faces together only by generating vertices at coarsely-discretized categorical coordinates.
However, by building on proven paradigms from language modeling, these models have been successfully trained at very large scale.
Additionally, many of these approaches support only unconditional generation, and some are not publicly available.
We include a gallery of qualitative comparisons in \figref{shapenet_qual}.

\subsection{Graph Learning}
Our approach draws inspiration from graph learning representations, which have shown success for graphs including gene expression \cite{marbach2012wisdom}, molecules \cite{kwon2020compressed}, stochastic processes \cite{backhoff2020all}, and social networks  \cite{gehrke2003overview}. 
Based on the seminal work of \citet{gromov1987hyperbolic},  \citet{nickel2017poincare} showed that hyperbolic embedding has fundamental properties which Euclidean embedding lacks (a relationship which has been well-studied in physics \cite{kronheimer1967structure,bombelli1987space,meyer1993spherical}), to exploit the geometry of \emph{spacetime} to represent graphs.
In this paper, we leverage spacetime embeddings~\cite{law2020ultrahyperbolic,law2023spacetime} to put this perspective to work for generating meshes.

\section{Representation}
\label{sec:Representation}

We propose a continuous representation for the space of manifold polygonal meshes, which requires no constraints and is suitable for optimization and learning. %

\subsection{Background}
\label{sec:Background}
\paragraph{Manifold Surface Meshes}

A surface mesh $\Mesh = (\Vertices,\Edges,\Faces)$ consists of vertices $\Vertices$, edges $\Edges$, and faces $\Faces$, where each vertex $\Vertex \in \Vertices$ has a position $\VertexPosition_\Vertex \in \RR^3$. 
In a general polygonal mesh, each face is a cyclic ordering of 3 or more vertices.
Each edge is an unordered pair of vertices which appear consecutively in one or more faces.

We are especially concerned with generating meshes which are not just a soup of faces, but which have coherent and consistent neighborhood connectivity.
As such, we consider \emph{manifold}, \emph{oriented} meshes.
Manifold connectivity is a topological property which does not depend on the vertex positions: \emph{edge-manifoldness} means each edge has exactly two incident faces, while \emph{vertex-manifoldness} means the faces incident on the vertex form a single edge-connected component homeomorphic to a disk.
In an \emph{oriented} mesh, all neighboring faces have a consistent outward orientation as defined by a counter-clockwise ordering of their vertices.

\setlength{\columnsep}{1em}
\setlength{\intextsep}{0em}
\begin{wrapfigure}{r}{80pt}
  \includegraphics{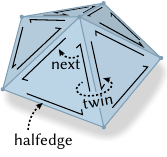}
\end{wrapfigure}

\paragraph{Halfedge Meshes}

There are many possible data structures for mesh connectivity; we will leverage \emph{halfedge meshes}, which by-construction encode manifold, oriented meshes with possibly polygonal faces, all using only a pair of references per element.
As the name suggests, halfedge meshes are defined in terms of directed face-sides, called \emph{halfedges} (see inset).
Each halfedge stores two references: a $\hetwin$ halfedge, the oppositely-oriented halfedge along the same edge in a neighboring face, and a $\henext$ halfedge, the subsequent halfedge within the same face.

\setlength{\columnsep}{0.5em}
\setlength{\intextsep}{0em}
\begin{wrapfigure}{l}{71pt}
  \includegraphics{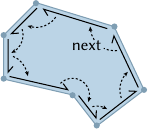}
\end{wrapfigure}
The $\hetwin$ and $\henext$ operators can be interpreted as a pair of permutations over the set of halfedges, this group-theoretic perspective is studied in combinatorics as a \emph{rotation system}.
A pair of permutations can be interpreted as a halfedge mesh as long as (a) neither operator maps any halfedge to itself, and (b) $\hetwin$ operator is an involution, \ie{} $\hetwin(\hetwin(\Halfedge)) = \Halfedge$.
The faces of the mesh are the \emph{orbits} traversed by repeatedly following the $\henext$ operator (see inset); we further require that these orbits all have a degree of at least three, to disallow two-sided faces.
Our representation will construct a valid set of $\hetwin$ and $\henext$ operators from a continuous embedding to define mesh connectivity.

\subsection{Representing Edges} 
\label{sec:RepresentingEdges}

To begin, consider modeling a mesh simply as a graph $\Graph = (\Vertices, \Edges)$, later we will extend this model to capture manifold mesh structure via halfedge connectivity (\secref{RepresentingFaces}).
The vertex set $\Vertices$ can be viewed as a particular kind of point cloud, and point cloud generation is a well-studied problem (\cite{nichol2022point,zeng2022lion}).
Likewise, continuous representations for generating undirected graph edges is a classic topic in graph representation learning \cite{nickel2017poincare, law2020ultrahyperbolic}.
A basic approach is to associate an \emph{adjacency embedding} $\AdjacencyFeature_\Vertex \in \RR^\AdjacencyFeatureDim$ with each vertex, then define an edge between two vertices $i,j$ if they are sufficiently close w.r.t. some distance function $\Distance$:
\begin{equation}
  \Edges := \big\{ \{i,j\} \textrm{ such that $\Distance(\AdjacencyFeature_i, \AdjacencyFeature_j) < \AdjacencyThreshold$}  \big\}
\end{equation}
for some learned threshold $\AdjacencyThreshold \in \RR$.
Representing the vertices and edges of a mesh then amounts to two vectors for each vertex $\Vertex$: a 3D position $\VertexPosition_\Vertex \in \RR^3$ and an adjacency embedding $\AdjacencyFeature_\Vertex \in \RR^\AdjacencyFeatureDim$.

\paragraph{Spacetime Distance}
We find that taking the adjacency features $\AdjacencyFeature$ as Euclidean vectors under pairwise Euclidean distance $\DistanceEuclidean
(\AdjacencyFeature_i, \AdjacencyFeature_j) = ||\AdjacencyFeature_i - \AdjacencyFeature_j||_2$ is ineffective, with poor convergence in optimization and learning.
There are many other possible choices of distance function for this embedding, but we find the recently proposed \emph{spacetime} distance \cite{law2023spacetime} to be simple and highly effective.
This distance function has deep interpretations in special relativity, defining pseudo-Riemannian structures.
In our setting the \emph{spacetime} distance $\DistanceSpacetime$ is computationally straightforward, splitting the components of $\AdjacencyFeature$ into a subvector $\AdjacencyFeatureSpacelike \in \RR^\AdjacencyFeatureSpacelikeDim$ of space coordinates, and a subvector $\AdjacencyFeatureTimelike \in \RR^\AdjacencyFeatureTimelikeDim$ of time coordinates:
\begin{equation}
  \label{eq:SpacetimeDistance}
  \DistanceSpacetime(\AdjacencyFeature_i, \AdjacencyFeature_j) = \DistanceSpacetime( [\AdjacencyFeatureSpacelike_i, \AdjacencyFeatureTimelike_i], [\AdjacencyFeatureSpacelike_j, \AdjacencyFeatureTimelike_j]) := 
  ||\AdjacencyFeatureSpacelike_i - \AdjacencyFeatureSpacelike_j||_2^2 - ||\AdjacencyFeatureTimelike_i - \AdjacencyFeatureTimelike_j||_2^2,
\end{equation}
where $[\cdot, \cdot]$ denotes vector concatenation.
Note that $\DistanceSpacetime$ is not a distance metric, and may be negative; this is of no concern, as we simply need to threshold it by some $\AdjacencyThreshold \in \RR$ to recover edges, treating $\AdjacencyThreshold$ as an additional optimized parameter.
In \figref{bridge_plot} we show that this significantly accelerates convergence, see \secref{verification} for details. 

\paragraph{Loss Function} At training time, we fit the adjacency embedding by supervising the distances under a cross entropy loss:
\begin{equation} 
\label{eq:loss_edge}
    \sum_{i,j \in \Edges_\textrm{gt}}\! \log \big(\sigma(\Distance(\AdjacencyFeature_i, \AdjacencyFeature_j) - \AdjacencyThreshold)\big)
    + \lambda
    \sum_{i,j \not\in \Edges_\textrm{gt}}\! \log \big(\sigma(\AdjacencyThreshold - \Distance(\AdjacencyFeature_i, \AdjacencyFeature_j))\big)
\end{equation}
where $\sigma$ is the logistic function (\ie~ a sigmoid), $\Edges_\textrm{gt}$ denotes the set of edges in the ground truth mesh, and $\lambda > 0$ is a regularization parameter balancing positive and negative matches. 

\subsection{Representing Faces} 
\label{sec:RepresentingFaces} 

To recover faces and manifold connectivity from a graph $\Graph = (\Vertices, \Edges)$, we further propose to parameterize halfedge connectivity for the mesh (\secref{Background}).
Given $\Vertices$ and $\Edges$, we construct the halfedge set by splitting each edge $e_{ij}$ between vertices $i$, $j$ into two oppositely-directed halfedges $\Halfedge_{ij}$,$\Halfedge_{ji}$.
This pairing trivially implies the $\hetwin$ relationships as $\hetwin(\Halfedge_{ij}) = \Halfedge_{ji}$; we then only need to specify the $\henext$ relationships to complete the halfedge mesh and define the face set.

\setlength{\columnsep}{1em}
\setlength{\intextsep}{0em}
\begin{wrapfigure}{r}{112pt}
  \includegraphics{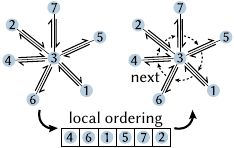}
\end{wrapfigure}
\paragraph{Neighborhood Orderings}
The $\henext$ operator defines a cyclic permutation with a single orbit on the halfedges outgoing from each vertex.
Thus the task of assigning the $\henext$ operator (and implicitly, the potentially-polygonal faces of the mesh) comes down to learning this permutation for each vertex.

\paragraph{Representing Neighborhood Orderings}
For each vertex, we define a triplet of continuous \emph{permutation features}: $\PermutationFeatureRoot, \PermutationFeaturePrev, \PermutationFeatureNext \in \RR^\PermutationFeatureDim$. These are used to determine the local cyclic ordering of incident edges.
Precisely, in the local neighborhood of each vertex $i \in \Vertices$ with degree $\VertexDegree$, for each pair of edges $\Edge_{ij}$,$\Edge_{ik}$, we combine the features of vertices $i,j$ and $k$ via a scalar-valued function $\PermutationReduction(\PermutationFeatureRoot_i, \PermutationFeaturePrev_j, \PermutationFeatureNext_k)$ (see \secref{design_choices_ablation}).
Gathering these pairwise entries yields a nonnegative matrix in the local neighborhood of each vertex:%
\begin{equation}
  \PermutationMatrix^i \in \RR^{\VertexDegree \times \VertexDegree},
  \qquad
  \PermutationMatrix^i_{jk} := e^{\PermutationReduction(\PermutationFeatureRoot_i, \PermutationFeaturePrev_j, \PermutationFeatureNext_k)},
\end{equation}
where each row corresponds to an incident edge.
We then use Sinkhorn normalization \cite{sinkhorn1964relationship} to recover a doubly-stochastic matrix, $\NormalizedPermutationMatrix^i$, representing a \emph{softened} permutation matrix~\cite{adams2011ranking}.

\paragraph{Loss Function}
At training or optimization time, we simply supervise the matrices $\NormalizedPermutationMatrix$ directly with the ground truth permutation matrices using binary cross-entropy loss:
\begin{equation}
\label{eq:loss_face}
\sum_{\{i,j,k\}\in \Next_\textrm{gt}} - \log(\NormalizedPermutationMatrix^i_{jk}),
\end{equation}
where $\Next_\textrm{gt}$ is the set of all $\henext$ relationships in ground truth mesh such that $\henext(\Halfedge_{ij}) = \Halfedge_{jk}$. Note that we do not need to supervise the remaining entries of $\NormalizedPermutationMatrix^i$, which is already Sinkhorn-normalized.

\paragraph{Extracting Meshes}
At inference time to actually extract a mesh, for each vertex neighborhood we seek the lowest-cost matching under the pairwise cost matrix $-\NormalizedPermutationMatrix^i$, among only those matchings which form a single orbit.
To compute this matching, we first compute the optimal unconstrained lowest-cost matching \cite{jonker1988shortest}; often this matching already forms a single orbit, but when it does not we fall back on a greedy algorithm which starts at an arbitrary entry and repeatedly takes the next lowest-cost entry without violating the single-orbit constraint.
These neighborhood matchings then imply halfedge connectivity as
\begin{equation}
  \henext(\Halfedge_{ij}) := \Halfedge_{ki} 
  \quad \textrm{for} \quad 
  k = \textrm{match}_{\PermutationMatrix^i}(j).
\end{equation}

This completes the halfedge mesh representation.
Faces, potentially of any polygonal degree, can then be extracted as orbits of the $\henext$ operator.

\section{Validation}
\label{sec:verification}

In this section, we evaluate the basic properties of our method, by directly optimizing to fit both individual meshes and collections of meshes, as well as ablating design choices.

\begin{figure}[t]
    \centering
    \includegraphics{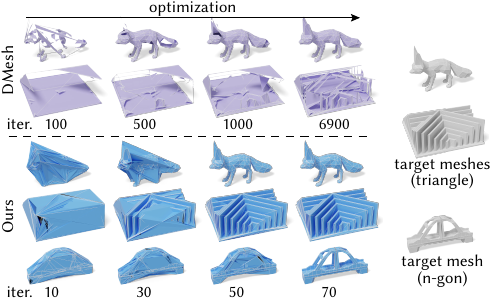}
    \caption{
        Fitting the ground truth connectivity of a single mesh tessellated with triangles and n-gons. 
    }
    \label{fig:mesh_to_rep}
\end{figure}

\subsection{Encoding a Given Mesh}
\label{sec:mesh_to_ours}
The most basic task for a mesh representation is to directly fit it to encode a particular mesh.
Though straightforward in principle, this optimization could fail if a representation is unable to represent all possible meshes, or if local minima and slow convergence make fitting ineffective in practice.
We consider three different challenging meshes with thin parts, anisotropic faces, and varying geometric details.
For each single shape, we optimize to encode its connectivity with our per-vertex embeddings $(\AdjacencyFeature_i,\PermutationFeatureRoot_i, \PermutationFeaturePrev_i, \PermutationFeatureNext_i)$ 
using the loss functions from ~\eqref{loss_edge} and ~\ref{eq:loss_face}.
In ~\figref{mesh_to_rep} we show the result of this optimization with our approach, as well as with the recent DMesh~\cite{son2024dmesh}, which proposes a Delaunay-based mesh representation.
Our method not only converges much faster to the correct connectivity, but also is applicable to polygonal meshes, making it more suitable for general mesh generation tasks.
Further experimental details are provided in the Supplement. 

\begin{figure}[b]
    \centering
    \includegraphics[width=0.98\columnwidth]{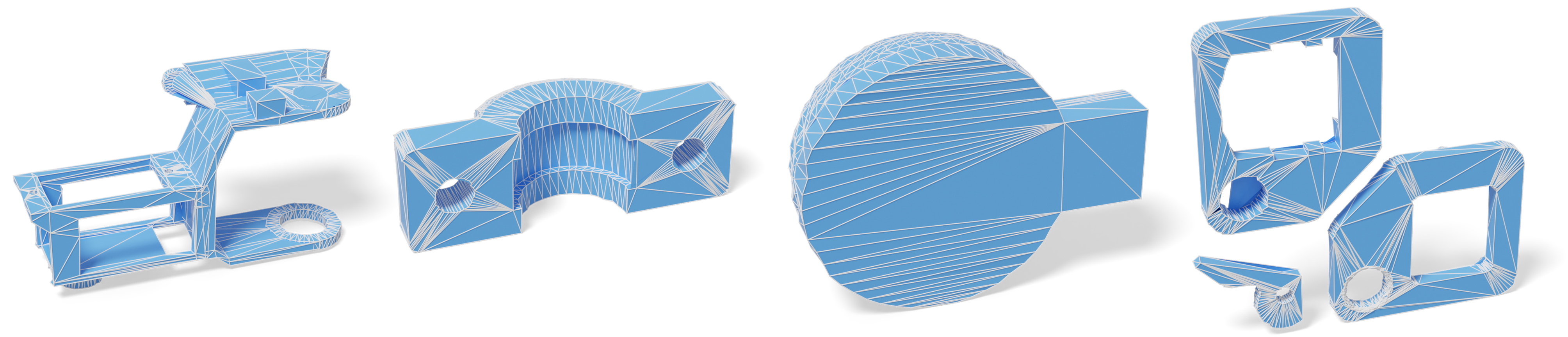}
    \caption{
        Meshes encoded by our autodecoder.
    }
    \label{fig:autodecoder_qual}
\end{figure}

\subsection{Fitting Mesh Collections}

As a next basic test of the ability of our method to encode collections of shapes in a learning setting, we train a simple auto-decoder architecture on a subset of 200 shapes from the Thingi10k dataset, a challenging set of real-world models originally for 3D printing~\cite{zhou2016thingi10k}.
To be clear, we do not aim to demonstrate downstream learning tasks with this experiment, we simply validate that our representation can simultaneously represent a variety of complex shapes, even when the embeddings are parameterized by a neural network, see \secref{application} for large-scale learning and applications.
In particular, here we allocate a latent code for each mesh, and optimize those latent codes as well as the parameters of a simple transformer model~\cite{vaswani2017attention} that decodes each latent code into the mesh, in the form of per-vertex positions and connectivity embeddings of our representation.
See the Supplement for further experimental details. 
As shown in \figref{autodecoder_qual}, our model faithfully overfits the shape collection.
Quantitatively, the encoded meshes achieve a mean L2 loss of 0.00062, an F1 score of 0.99 for adjacency prediction, and an accuracy of 0.98 for permutation predictions. 
This is positive evidence that the representation is able to  simultaneously represent many complex shapes, even with significant geometric complexity and the nonconvexity of the neural parameterization.

\begin{figure}[t]
    \centering
    \includegraphics{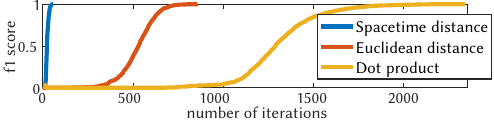}
    \captionsetup{skip=4pt}
    \caption{
        Comparison of convergence speed with different distance functions. 
        \label{fig:bridge_plot}
    }
\end{figure}

\begin{figure}
    \centering
    \includegraphics{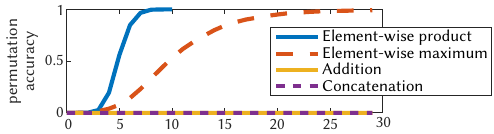}
    \captionsetup{skip=4pt}
    \caption{
            Comparison of convergence speed with different permutation feature reduction functions. 
    \label{fig:bridge_permutation}
    }
\end{figure}

\subsection{Ablating Design Choices}
\label{sec:design_choices_ablation}

\paragraph{Spacetime Distance} \label{sec:ablation_spacetime_distance}
We find spacetime distance to be a dramatically more effective representation than Euclidean or other metrics to define  adjacency embeddings (\secref{RepresentingEdges}), in the sense that it can be optimized much more easily.
To demonstrate this, we fit the edges of the bridge mesh appearing on the bottom right of ~\figref{bridge_plot} using each of three formulations for $\dsf(\xvec_i, \xvec_j)$: (1) the spacetime distance introduced in Section \ref{sec:RepresentingEdges}, (2) the squared Euclidean distance $\| \xvec_i - \xvec_j \|_2^2$, and (3) the negative dot product  $- \xvec_i^{\top} \xvec_j$. 
\figref{bridge_plot} shows the speed of convergence---spacetime distance converges much faster compared to other distance formulations, which we observed consistently across all experiments.

\paragraph{Permutation Feature Reduction}
We also investigate several choices for the permutation feature reduction function $\PermutationReduction$ (\eqref{perm_reduction}), including elementwise maximum, addition, or concatenation.
\figref{bridge_permutation} shows the results. 
We find elementwise multiplication followed by summation of all elements to be most effective.
Precisely, we use 
\begin{equation} 
  \label{eq:perm_reduction}
  \PermutationReduction(\PermutationFeatureRoot_i, \PermutationFeaturePrev_j, \PermutationFeatureNext_k) := 
  \textrm{trace}\big( \textrm{diag}(\PermutationFeaturePrev_j) \textrm{diag}(\PermutationFeatureRoot_i) \textrm{diag}(\PermutationFeatureNext_k) \big),
\end{equation} 
where $\textrm{diag}$ denotes constructing a diagonal matrix from a vector.

\section{Application: Learning to mesh}
\label{sec:application}
Equipped with a continuous representation for manifold polygonal meshes, we can then begin large-scale learning atop the representation.
In this section, we integrate \ours{} with a 3D generative model to generate meshes conditioned on geometry provided as a point cloud. This conditioned model can then be directly applied to mesh repair without fine-tuning (\secref{mesh_repair}).

\begin{figure}[b]
    \centering
    \includegraphics{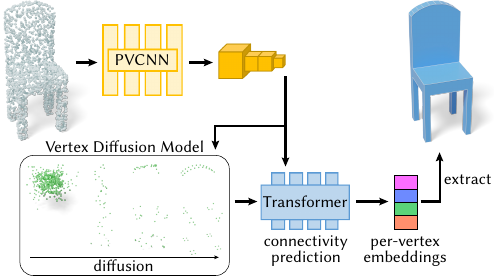}
    \caption{
        Network architecture for learning to generate meshes.
    }
    \label{fig:arch}
\end{figure}

\subsection{Model Architecture}
Our model architecture (Figure~\ref{fig:arch}), consists of three modules: a point cloud encoding network for processing geometry information, a vertex diffusion model to generate 3D locations for vertices, and a connectivity prediction network to predict per-vertex embeddings. 

\paragraph{Point Cloud Encoder} We encode the point cloud using PVCNN~\cite{liu2019point} to generate the feature volumes at multiple spatial resolutions. These feature volumes, as geometry context, guide the subsequent mesh generation. 
Note that this input point cloud is \emph{not} the resulting mesh vertex set, it is conditioning information indicating the geometry that we are trying to generate a mesh of.

\paragraph{Vertex Position Generation Network} We re-purpose Point-E~\cite{nichol2022point}, a diffusion transformer network that was originally designed for point cloud generation, to generate sparse mesh vertices conditioned on the geometry context from the encoder. Specifically, we first initialize the vertex position by sampling from a Gaussian distribution, and iteratively denoise the vertex location through the diffusion transformer. At each denoising step, we feed the input to the transformer by concatenating the vertices' positions with features that are tri-linearly interpolated with the multi-resolution feature volumes from the encoder to capture the geometry information.
If needed, we handle varying vertex counts by padding to a predefined maximum size, and additionally diffusing a binary mask at each vertex to indicate which vertices are artificial padding.

\paragraph{Vertex Connectivity Prediction Network}
We leverage a transformer architecture~\cite{vaswani2017attention} to predict the per-vertex connectivity embeddings given vertex positions. Similar to the vertex position generation network, we concatenate vertex position with the interpolated feature from the encoder for each vertex, and predict the adjacency embeddings $x$ and permutation embeddings $\PermutationFeatureRoot, \PermutationFeaturePrev, \PermutationFeatureNext$. We remove the positional embedding from the original transformer and predict the embeddings for all the vertices simultaneously by using the self-attention across the vertices.

\paragraph{Training Details}
We train all the neural networks together. To train the vertex position generation network, we adopt the $\epsilon$-prediction from the  diffusion model~\cite{ddpm,nichol2022point}. To train the connectivity generation model, we combine the losses \eqref{loss_edge} and \ref{eq:loss_face}, supervising on meshes from the dataset.
Further details are provided in the Supplement.

\begin{figure}[t]
    \centering
    \includegraphics{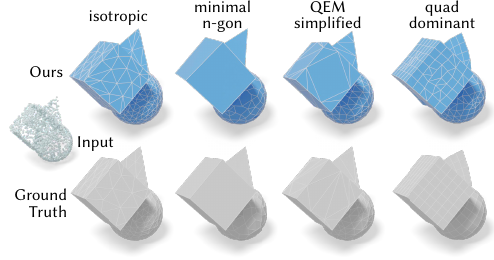}
    \caption{
        Conditioned on the same geometry, our model can generate different styles of meshes depending to the distribution it was trained on.
        Each row denotes a style of mesh, for which we construct a dataset of meshed primitive surfaces and fit our model.
        Because our model is generative, it matches the distribution but does not exactly replicate vertex positions or connectivity.
    }
    \label{fig:toy_exp}
\end{figure}

\subsection{Basic Validation on a Synthetic Dataset}
Our model learns to fit \emph{distributions} of meshes; the tessellation pattern and element shapes of generated meshes will mimic the training population.
We first demonstrate this behavior with a simple synthetic dataset, constructed by generating shapes as a union of randomly arranged cubes, tetrahedra, and spheres.
For each shape, we extract a 3D iso-surface using Dual Marching Cubes~\cite{nielson2003marching}, and mesh it according to several strategies:
(1) isotropic remeshing~\cite{hoppe1993mesh} with Meshlab~\cite{cignoni2008meshlab}
(2)  planar decimation from Blender~\cite{blender} to create N-gon mesh.
(3) QEM for surface simplification~\cite{garland1997surface} from Meshlab,
and 
(4) InstantMesh~\cite{Jakob2015Instant} to create a quad-dominant mesh with the official implementation

In ~\figref{toy_exp}, we show how training on each of these datasets causes our model to generate different styles of meshes as outputs.
The four models, when each given the same point cloud as input specifying the desired geometry, produce respectively (1) isotropic triangle meshes, (2) minimal planar-decimated meshes, (3) QEM-simplified meshes, and (4) quad-dominant meshes.

\subsection{Learning Meshes from the ABC Dataset}
\label{sec:exp_abc}
To evaluate learning at scale on a realistic dataset,  train our model on ABC dataset~\cite{Koch_2019_CVPR}, which consists of watertight triangle meshes of CAD shapes with isotropic triangle distribution. The meshes in the ABC dataset exhibit considerable diversity, featuring both sharp and smooth curved geometric features. We employed a benchmark~\cite{koch2019abc} subset of 10,000 shapes, all with 512 vertices, randomly split into 80\% for training and 20\% for testing. To obtain the input conditioning point cloud, we uniformly sampled 2048 points from the mesh surface.

\paragraph{Baselines}
We compare our model against both classic and learning-based point cloud reconstruction methods.
As a representative classic approach, we compare to Poisson Surface Reconstruction (PSR)~\cite{kazhdan2006poisson} as implemented in Open3D~\cite{zhou2018open3d}, with meshes extracted via marching cubes~\cite{lorensen1998marching}. 
We also consider isotropic remeshing~\cite{hoppe1993mesh} on the output of Poisson reconstruction to obtain a more compact mesh tessellation, which is denoted PSR$^*$. 
For representative learning-based approaches, we choose Pixel2Mesh~\cite{Wang2018Pix2Mesh}, which deforms a template sphere to generate a mesh, and OccNet~\cite{occnet}, which predicts an implicit field and extracts the mesh using Marching Cubes~\cite{lorensen1998marching} afterwards. For a fair comparison among deep learning based methods, we adopt the same point cloud encoder as our approach.

\begin{table}[]
\caption{Accuracy and quality statistics for mesh reconstruction.}
\resizebox{\columnwidth}{!}{
\begin{tabular}{l|ccccccc}
\hline
Method & CD $(10^{-3})\downarrow$ & F1$\uparrow$ & ECD$(10^{-2})\downarrow$ & EF1$\uparrow$ & \#V & \#F & IN$\downarrow$ \\ \hline
PSR & 46.35 & 0.44 & 56.81 & 0.03 & 2406 & 4736 & 63.31 \\
PSR$^*$ & 46.72 & 0.42 & 51.86 & 0.03 & 494 & 968 & 61.61 \\
OccNet & 11.31 & 0.47 & 33.08 & 0.08 & 7344 & 14688 & 48.53 \\
Pixel2Mesh & 6.37 & 0.48 & 29.52 & 0.09 & 2466 & 4928 & 52.03  \\
Ours & 1.39 & 0.66 & 3.21 & 0.42 & 512 & 1818 & 34.54 \\ \hline
\end{tabular}
}
\label{tbl:abc_quan}
\end{table}

\begin{figure*}[t]
    \centering
    \includegraphics[width=0.8\linewidth]{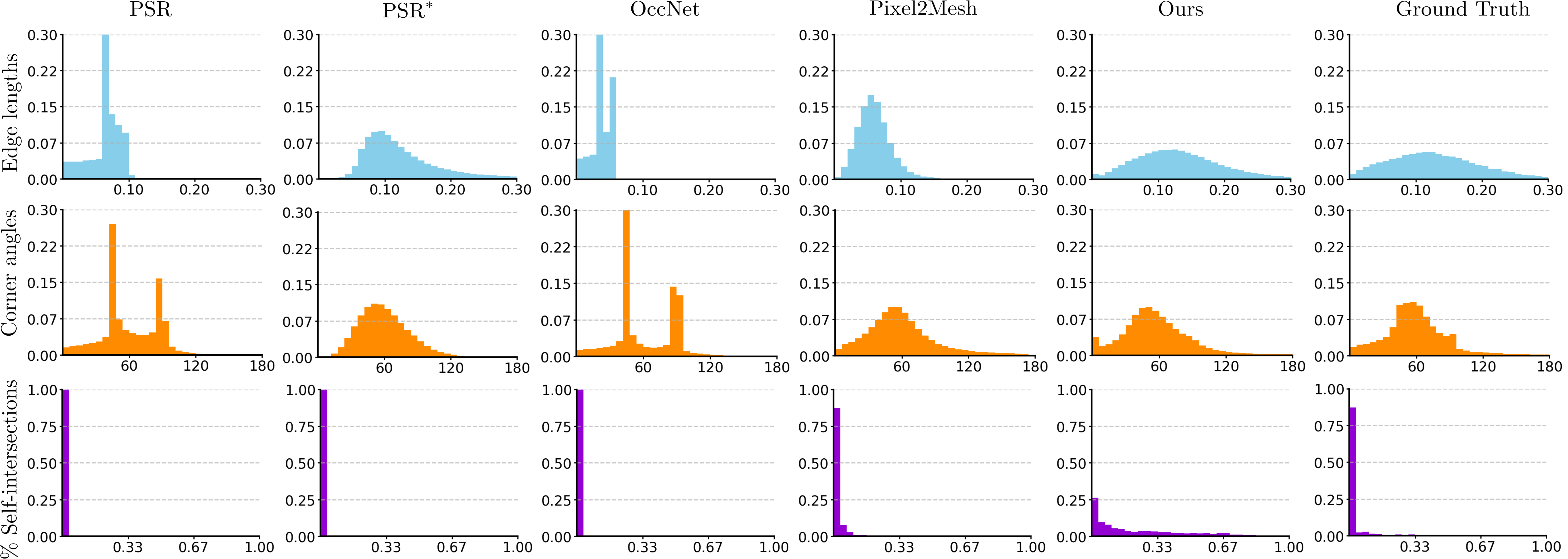}
    \captionsetup{skip=4pt}
    \caption{
        Quantitative comparison of the intrinsic quality of reconstructed meshes. Our method produces meshes with distributions of edge lengths and corner angles more closely aligned with the ground truth. This is because our model learns the surface discretization from the data, unlike other methods that primarily focus on reconstructing geometry. We additionally report the percentage of faces with self-intersections in each mesh.
    }
    
    \label{fig:abc_histogram}
\end{figure*}

\paragraph{Metrics}
Our primary goal is to evaluate the ability to capture the desired distribution of surface discretization, as measured by intrinsic mesh statistics such as edge lengths and corner angles for each polygon. 
Furthermore, although our method is not directly designed to minimize reconstruction error, we additionally evaluate our method against baselines on how well the generated meshes align with ground truth geometry. To this end, we follow the methodology from NDC~\cite{chen2022ndc} and compute Chamfer Distance (CD), F-Score (F1), Edge Chamfer Distance (ECD), Edge F-Score (EF1), and the percentage of Inaccurate Normals (IN> $10^\circ$) with respect to the ground truth mesh. A detailed description of these metrics is provided in the Supplement.

\paragraph{Results}
As shown in Figure~\ref{fig:abc_qual} and Table~\ref{tbl:abc_quan}, both qualitative and quantitative results demonstrate that our method outperforms baselines under the target metrics, particularly in recovering sharp features. 
The vertices and edges align accurately with sharp features, highlighting the advantage of directly generating meshes as the output representation. As shown in Figure~\ref{fig:abc_histogram}, the distribution of element shapes from our generated meshes aligns much better with the ground truth than the baselines, demonstrating the ability of our model to predict connectivity which aligns with the target training population. Note that although our representation guarantees manifold connectivity, there may still be geometric self-intersections between faces. We report the fraction of faces in each mesh with self-intersections in \figref{abc_histogram}, and provide further discussion in Section~\ref{sec:limitations_and_future_work}.

\begin{figure}
    \includegraphics{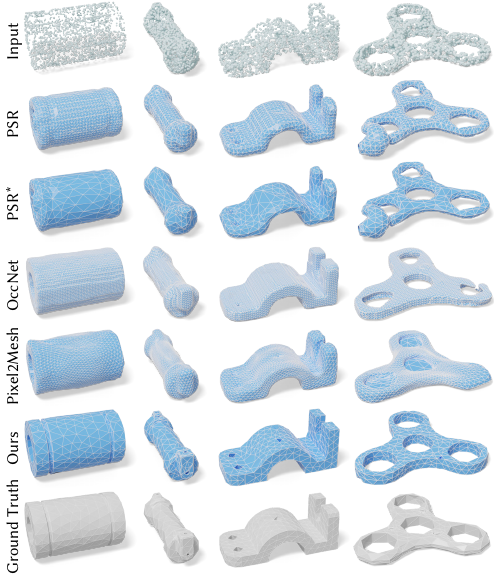}
    \captionsetup{skip=4pt}
    \caption{Generated meshes for the ABC Dataset.}
    \label{fig:abc_qual}
\end{figure}
\begin{figure}
    \centering
    \includegraphics[width=0.98\columnwidth]{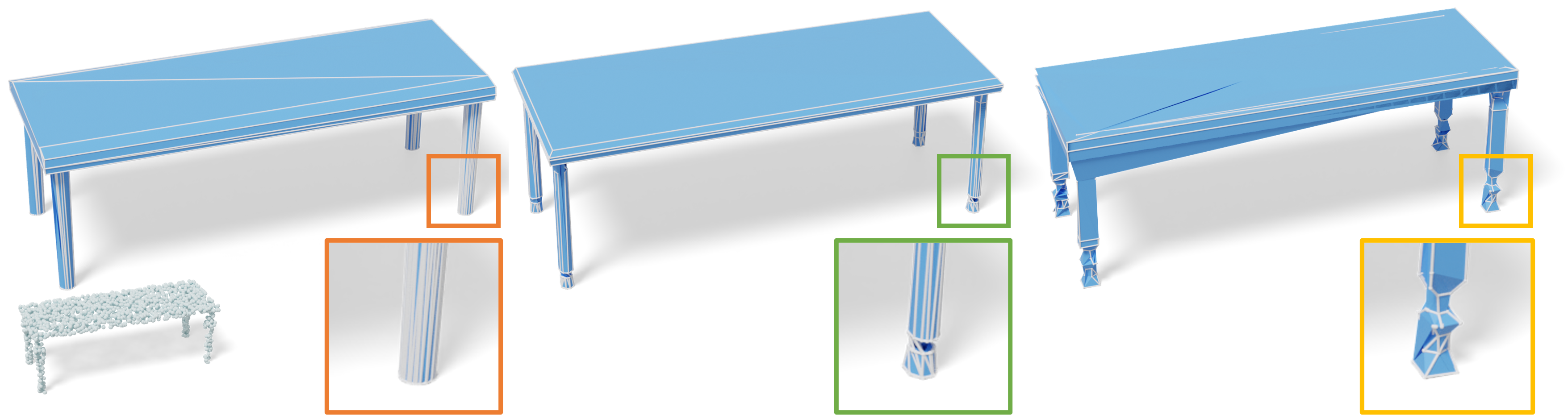}
    \caption{
        By leveraging a diffusion model, we can generate different meshes from the same input condition. Notice how the chair legs are modeled with different topologies, which all conform to the input condition.
    }
    \label{fig:shapenet_generative}
\end{figure}

\subsection{Learning Meshes from the ShapeNet Dataset}
Following recent work on mesh generation~\cite{siddiqui2023meshgpt,nash2020polygen,alliegro2023polydiff,gao2022get3d}, we further evaluate our model on ShapeNet dataset~\cite{shapenet}. 

\paragraph{Dataset Details} 
As in prior work~\cite{nash2020polygen,siddiqui2023meshgpt}, we note that the raw meshes from ShapeNet consist largely of non-manifold meshes with duplicated faces and T-junctions at intersections, and thus we preprocess all shapes by removing duplicated faces and applying planar decimation with varying thresholds to simplify them into minimal polygonal meshes. 
After this preprocessing, the majority of the shape are still non-manifold, making them unsuitable for our goal of generating manifold meshes with clean connectivity. We thus remove all non-manifold shapes, resulting in a total of 20,255 shapes. We adhere to an 80-20 train-test split and randomly sample 2,048 surface points as geometry conditioning input. 
Additionally, we apply random scaling augmentation during training. 
Unlike previous autoregressive methods, our approach does not require quantization of vertices.

\paragraph{Baselines} 
Many relevant baselines ~\cite{nash2020polygen,siddiqui2023meshgpt,alliegro2023polydiff} do not have either training or inference code available, and regardless there are many differences in experimental protocols and target task.
As such, we instead focus on qualitative comparisons to give intuition about the differences between these methods, primarily in regard to mesh quality.

\begin{figure}
    \centering
    \includegraphics{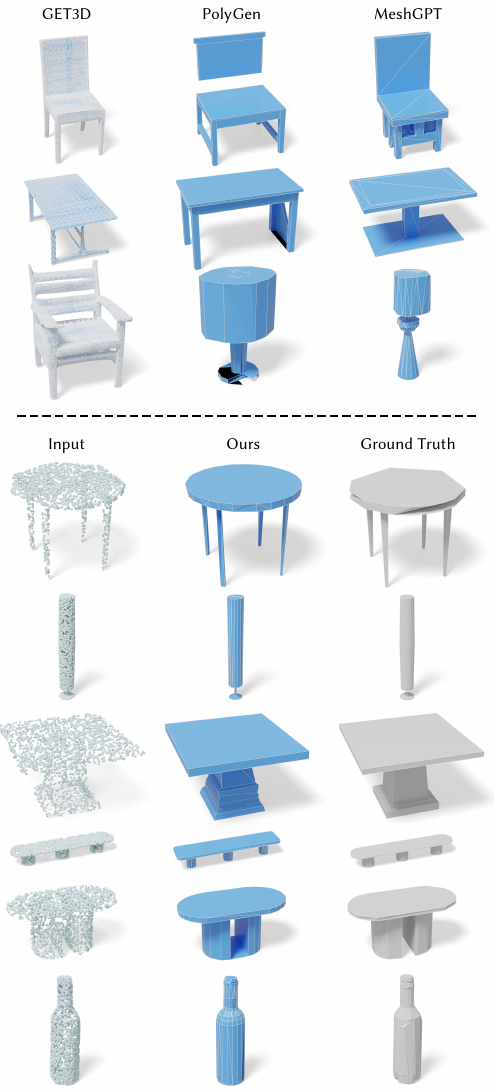}
    \caption{
        Visual comparison of generated meshes from models trained on ShapeNet. The top part showcases results from three unconditioned mesh generation methods: GET3D, PolyGen, and MeshGPT. The bottom part shows meshes generated by our model, which takes point clouds as input.
    }
    \label{fig:shapenet_qual}
\end{figure}

\paragraph{Results}
\figref{shapenet_qual} shows a variety of results generated by our method, as well as a sampling of published results from baselines.
Our method generates sharp and compact polygonal meshes that match with the input condition and are guaranteed to be manifold. %
We also note a promising \emph{diversity} in our outputs on this dataset: because our model uses a probabilistic diffusion model to generate vertices, we are able to produce distinct meshes conditioned on the same point cloud input by repeatedly sampling the model (Figure~\ref{fig:shapenet_generative}).

\begin{figure}
    \centering
    \includegraphics[width=0.98\columnwidth]{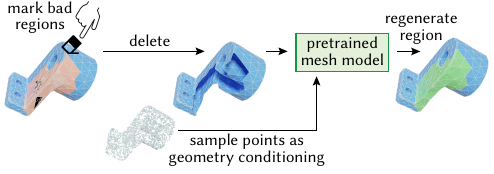}
    \caption{
        Our trained conditioned meshing model can be repurposed for mesh repair. For a mesh with good geometry but poor tessellation in certain regions (highlighted in red), the user can mark those regions and pass the mesh to our model to re-predict both vertices and connectivity, effectively repairing the mesh (highlighted in green).
    }
    \label{fig:mesh_repairing}
\end{figure}

\begin{figure}
    \centering
    \includegraphics{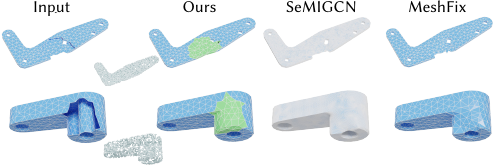} 
    \caption{A visual comparison of mesh repair methods. Note that our method additionally takes surface points sampled from the whole mesh as input, unlike other methods which use only the partial mesh.}
    \label{fig:abc_inpainting}
\end{figure}

\subsection{Mesh Repair}
\label{sec:mesh_repair}

Lastly, we demonstrate the application of our model to the downstream geometry processing task of mesh repair.
As illustrated in Figure~\ref{fig:mesh_repairing}, we envision a workflow where a user identifies a region of a mesh with poor tessellation such as self-intersections, skinny triangles, or non-manifold structures, and wishes to re-triangulate that region in a way that seamlessly blends with the surrounding mesh.
We show that we can repurpose our model for this task without retraining, by viewing it as \emph{mesh inpainting}, in the same sense that image models are used to inpaint undesired regions of images according to some conditioning while matching the surrounding context.
We inpaint the mesh by sampling a point cloud from the desired geometry and applying our generative model, projecting during diffusion to ensure the fixed region of the input mesh is retained---see the Supplement for an in-depth explanation.
Note that MeshGPT~\cite{siddiqui2023meshgpt} also demonstrated completion of a partial mesh; however, it was limited to bottom-up completion due to auto-regressive inference with sorted vertices.

\paragraph{Results} 
We visualize the results in Figure~\ref{fig:abc_inpainting}. 
Our approach generates high-quality patches to fill the removed regions in the partial meshes while preserving the geometry and connectivity of the input. 
For comparison, we also include the most similar results of which we are aware: a classic mesh repair framework, MeshFix~\cite{attene2010lightweight}, and a recent learning-based method, SeMIGCN~\cite{hattori2024learning}.
However, note that this is not exactly an apples-to-apples comparison, our method additionally takes the surface point cloud of the complete shape as input, with a focus on re-generating surface discretization while preserving geometry.
MeshFix is designed only for hole filling and cannot generate a repaired mesh conditioned on the geometry. 
In contrast, SeMIGCN re-meshes the shape for running GCN, resulting in an overly dense mesh that might not be desirable. 
We compare quantitatively with 100 randomly sampled examples from the ABC dataset validation shapes. \ours{} achieved a Chamfer Distance (CD) of 0.77 ($10^{-3}$) and a 0.76 F1 score. The baselines, SeMIGCN and MeshFix, achieve a CD of 39.50 ($10^{-3}$) and 31.59 ($10^{-3}$), and an F1 score of 0.57 and 0.72, respectively.

\begin{figure}[t]
    \centering
    \includegraphics[width=0.8\columnwidth]{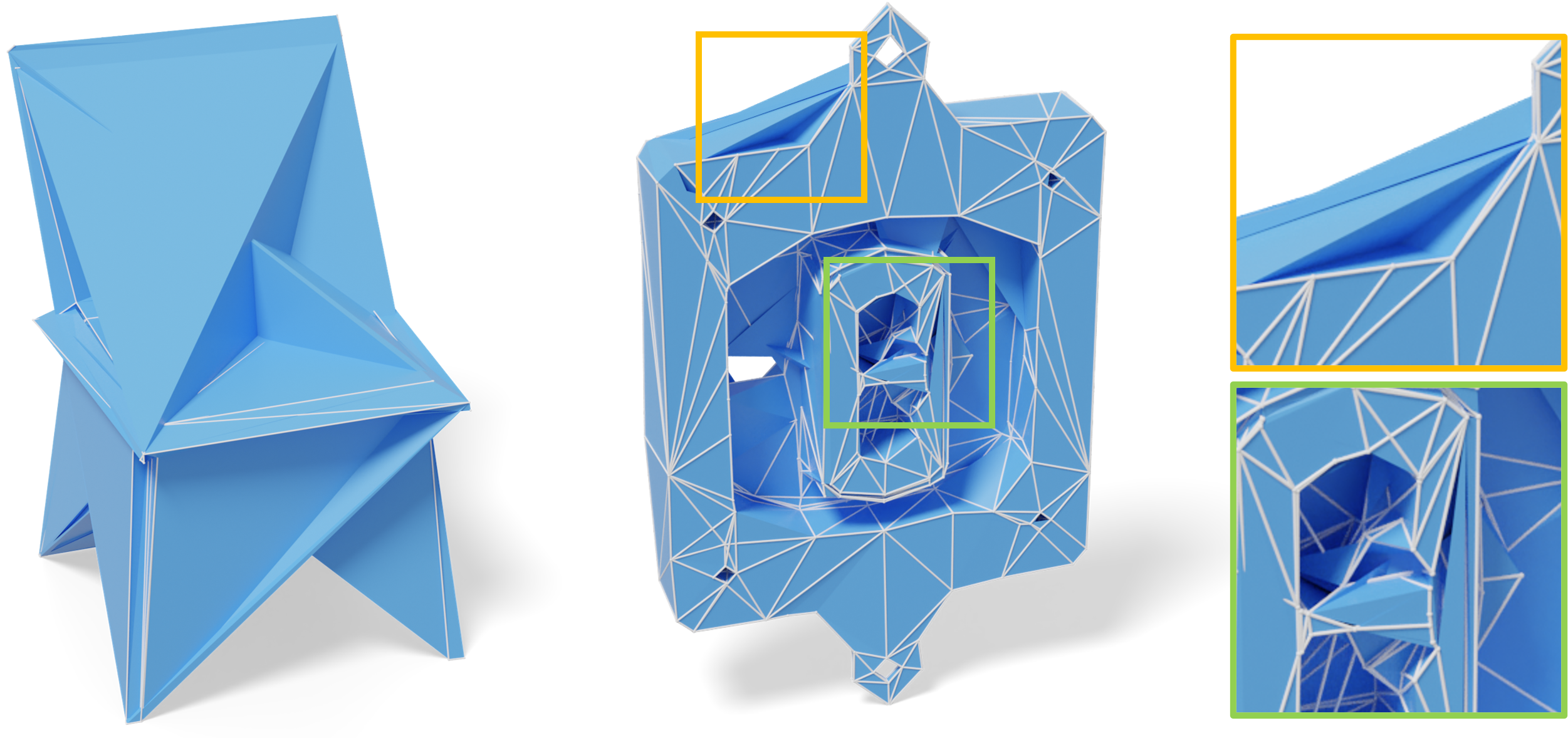}
    \caption{
        Failure cases from our learning to mesh model. Although they still have manifold connectivity, large erroneous faces and excessive self-intersections yield a tangled mesh with poor geometric accuracy.
    }
    \label{fig:limitation}
\end{figure}

\section{Discussion}
\label{sec:limitations_and_future_work}
\paragraph{Scalability and Runtime}
Our approach represents discrete connectivity via a fixed-size continuous embedding per-vertex.
Concrete results about the size of such an embedding needed to represent all possible discrete structures remain an open problem in graph theory~\cite{nickel2014reducing,nickel2017poincare}.
In practice, we find low-dimensional embeddings $k < 10$ to be sufficient to represent every mesh in our experiments. 
Encoding a 10,000-vertex mesh via direct optimization, as shown in \figref{mesh_to_rep}, converges in 600 iterations (approximately 2 minutes) with \( k^p = 6 \).

For learning, the bottleneck is memory usage in transformer blocks. 
We demonstrate generations up to 2,000 vertices in the auto-decoder setting; this is modest compared to high-resolution meshes, but it already captures many CAD and artist-created assets, and exceeds other recent direct mesh generation works (\eg{}, around 200 vertices in MeshGPT~\cite{siddiqui2023meshgpt}). Our generative model takes less than 2 seconds to generate a single mesh, which is notably faster than recent auto-regressive models like MeshGPT, which require 30-90 seconds. All inference and optimization times are measured on an NVIDIA A6000 GPU.

\paragraph{Limitations}
Although our representation guarantees manifold connectivity, it may contain other errors such as self-intersections, spurious high-degree polygons, or significantly non-planar faces. 
The frequency of such errors depends on how the representation is generated or optimized: often they have little effect on the approximated surface (\figref{abc_qual}), but in other cases they may significantly degrade the generated geometry, as shown in \figref{limitation}.
Note that such artifacts are not always erroneous---meshes designed by artists often intentionally include self-intersections; if desired, we could potentially mitigate self-intersections by penalizing them with regularizers during training.

Our implementation does not handle open surfaces, this could be addressed by predicting a flag for boundary edges much like we predict a mask for padded vertices.
Also, like other diffusion-based generative models, our large-scale learning experiments may produce nonsensical outputs for difficult or out-of-distribution input.

\paragraph{Future Work}
Looking forward, we see many possibilities to build upon our representation for directly generating meshes in learning pipelines.
In the short term, this could mean generating connectivity embeddings as well as vertex positions from a diffusion model, and in the longer term, one might even fit SpaceMesh generators in an unsupervised fashion using energy functions to remove the reliance on mesh datasets for supervision entirely.

\begin{acks}
  The authors are grateful to Yawar Siddiqui for providing the results of MeshGPT and Polygen, as well as the anonymous reviewers for their valuable comments and feedback.
\end{acks}

\clearpage
{\small
\bibliographystyle{ACM-Reference-Format}
\bibliography{mesh_generation_bib}
}

\clearpage
\appendix

In the supplement, we provide additional experimental details for mesh fitting (\secref{suppl_mesh_fitting} and 
\secref{suppl_mesh_collection_fitting}) and learning experiment (\secref{suppl_learning}). We further provide methodology details for applying our method to mesh repair tasks in \secref{suppl_mesh_repair}.

\section{Single Mesh Fitting }
\label{sec:suppl_mesh_fitting}
To fit a single mesh using our \ours~representation, we optimize the per-vertex embeddings $(\AdjacencyFeature_i, \PermutationFeatureRoot_i, \PermutationFeaturePrev_i, \PermutationFeatureNext_i)$. The adjacency embeddings, $\AdjacencyFeature_i$, are set to a dimension of 16, while each permutation embedding has a dimension of 6. The same dimensions for space and time coordinates are used across all experiments, with $k^s=k^t=k/2$. During training, the permutation matrix $\PermutationMatrix$ is constructed using the ground truth adjacency matrix. The regularization parameter $\lambda$ in Eq. 3 of the main paper is set to $\lambda = 4\frac{N_{\text{edges}}}{{N_{\text{vertices}}}^2}$, where $N_{\text{edges}}$ and $N_{\text{vertices}}$ represent the number of edges and vertices in each shape, respectively.

The Adam optimizer~\cite{kingma2014adam} is employed with a learning rate of 0.1. The overfitting process typically converges within 70 iterations. For comparison with DMesh~\cite{son2024dmesh}, we used the official released code\footnote{\url{https://github.com/SonSang/dmesh}}. For comparison with DSE~\cite{rakotosaona2021learning}, we used the official released code\footnote{\url{https://github.com/mrakotosaon/dse-meshing}} and trained the networks to overfit a single mesh each time.

\section{Fitting Collections of Meshes}
\label{sec:suppl_mesh_collection_fitting}
In this experiment we follow a typical autodecoder setup, optimizing one 512-dimension latent code for each mesh in the dataset and employing a Transformer~\cite{vaswani2017attention} to decode each latent code into the corresponding mesh. Specifically, for each mesh, the latent code is repeated $N$ times (where $N$ is the maximum number of vertices in the dataset), and a learnable positional embedding (shared across different meshes) is added to the repeated latent code. This resulting tensor is then passed to the Transformer to predict the vertex positions and per-vertex embeddings. To accommodate varying numbers of vertices across different meshes, all mesh vertices are padded with zeros up to $N=2000$.

Additionally, a mask channel is appended to the vertex positions, with ground truth values of -1 or 1 indicating whether the vertex is padded or not, respectively. During training, the latent code for each shape, the positional embedding, and the Transformer weights are jointly optimized using a combination of the L2 loss on the predicted vertices and the losses described in Section 3. During inference, vertices with negative predictions in the mask channel are pruned.

Consistent with the overfitting experiment, the regularization parameter $\lambda$ in Eq. 3 is set to $\lambda = 4\frac{N_{\text{edges}}}{{N_{\text{vertices}}}^2}$, and a learning rate of 0.001 is used.

\paragraph{Dataset}
We conduct our experiments on the Thingi10K~\cite{zhou2016thingi10k} dataset, which has a diverse collection of real-world 3D printing models exhibiting a variety of shape complexities, topologies, and discretizations. From this dataset, we filter a subset consisting of manifold meshes with vertex counts ranging between 1000 and 2000, and randomly select 200 meshes. For each selected mesh, the vertices are sorted in z-y-x order, in accordance with the methodology of PolyGen~\cite{nash2020polygen}.

\section{Learning to Mesh the Shape}
\label{sec:suppl_learning}
\subsection{Training Details}
In this experiment, we use a dimension of 32 for adjacency embeddings and a dimension of 12 for each permutation embedding. The total number of channels for the output of the vertex connectivity prediction network is 68.

For the point cloud encoder, we use a PVCNN~\cite{liu2019point} with 4 PVConv layers, each with voxel resolutions of 32, 16, 8, and channels of 64, 128, and 256, respectively. The vertex position generation network follows the transformer-based diffusion model as in the original Point-E~\cite{nichol2022point}, with 10 residual self-attention blocks of width equals 256. As mentioned in the main paper, at each denoising step, we concatenate the vertices' positions with features that are tri-linearly interpolated with the multi-resolution feature volumes from the encoder. The concatenated features are further passed through 4 PVConv layers with the same dimensions as the point cloud encoder. The training follows the standard diffusion model scheme with 1024 diffusion timesteps.

For the vertex connectivity prediction model, we also use PVConv to process the interpolated features. During training, our vertex generation model, which is based on a set transformer, does not hold correspondence between denoised vertices and ground truth vertices. Consequently, we use the ground truth vertices as input for training the connectivity prediction model, allowing supervision by the ground truth connectivity. We observe that the vertex generation model converges more quickly than the connectivity prediction network. Therefore, we train the connectivity prediction network for 10 steps for every single training step of the vertex generation model.

For the ABC dataset, since all meshes have the same number of vertices, we do not additionally predict a vertex mask. For the ShapeNet dataset, we predict 512 vertices, which is the maximum number of vertices in the filtered dataset, and an additional mask channel. We train our model using the Adam optimizer with a learning rate of 0.0001 for 800k iterations until convergence. To combine the loss for vertex prediction and connectivity prediction, we multiply the loss function from Equation. 3 by 200 and add it to the loss in Equation. 5 and the diffusion loss, both with a scale of 1.

\paragraph{Evaluation Metrics} 
We briefly describe the metrics used to evaluate the reconstruction quality. In all experiments, the longest dimension of all meshes is normalized to [-1, 1]. 

Chamfer Distance (CD) measures the distance between two point clouds using nearest neighbor search, sampling 10,000 points from the surface of each mesh. The F1-score is computed for the same point sets used for CD. Precision is determined by classifying points on the predicted mesh as true positives if their distance to the nearest ground truth (GT) point cloud is less than 0.02; otherwise, they are false positives. Recall uses the same 0.02 threshold. Edge Chamfer Distance (ECD) and Edge F-score (EF1) evaluate the reconstruction of sharp features, following prior works~\cite{chen2022ndc}. Each point in the sampled point cloud is checked by comparing the dot products between its normal and those of its neighbors; if the mean dot product is below 0.2, the point is classified as an edge point. ECD and EF1 then measure the Chamfer Distance and F1-score between these edge points. For the percentage of inaccurate normals, we use a 10-degree threshold. The normal of a point sampled from the reconstructed mesh is considered as inaccurate if the angles between its normals and that of the nearest ground truth point exceed 10 degrees.

\subsection{Additional Results}

\begin{figure}[t]
    \centering
    \includegraphics[width=0.98\columnwidth]{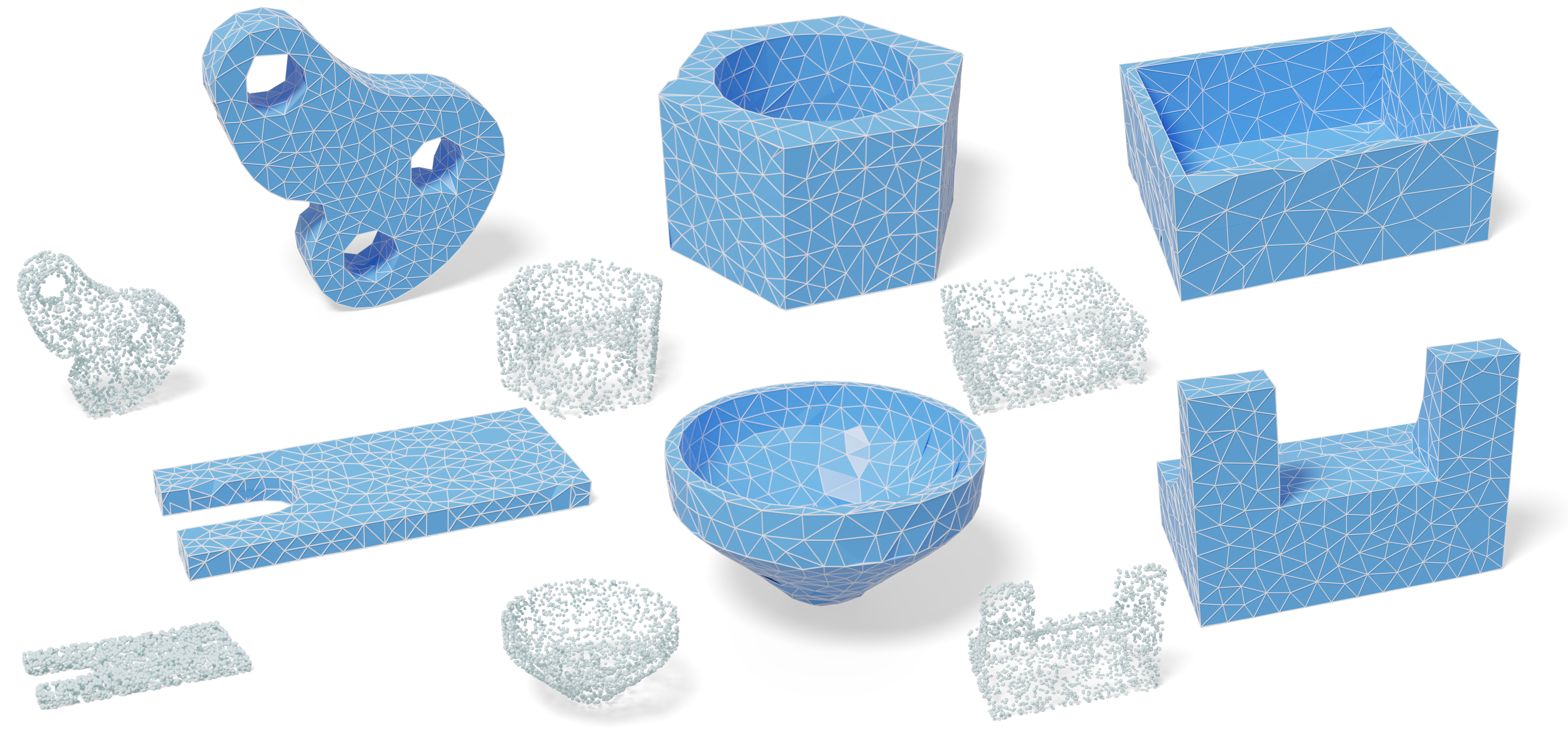}
    \caption{
        Results on the ABC dataset generated by our learning to mesh model.
    }
    \label{fig:gallery_abc}
\end{figure}

\begin{figure}[t]
    \centering
    \includegraphics[width=0.98\columnwidth]{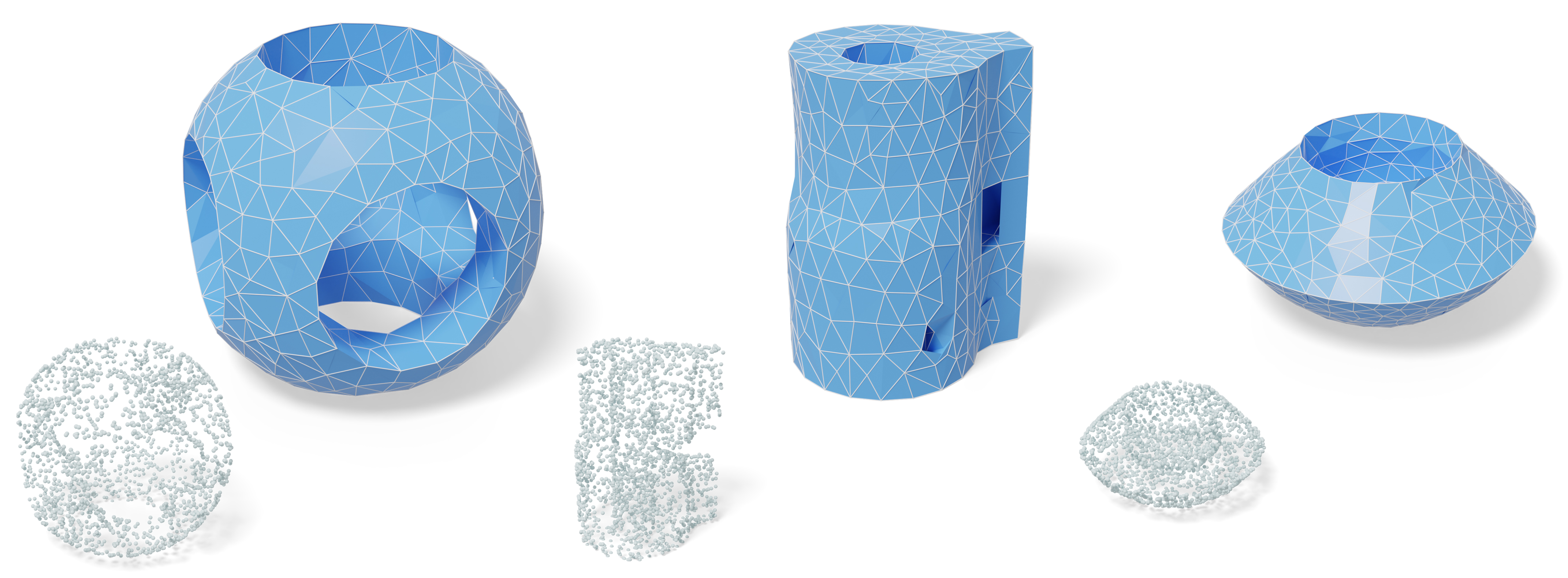}
    \caption{
        Results on the Thingi10k dataset generated by our learning to mesh model trained on ABC dataset.
    }
    \label{fig:abc_generalization}
\end{figure}

\paragraph{Results on ABC Dataset} 
We present additional qualitative results produced by our learning-to-mesh model, trained on the ABC dataset, in Figure~\ref{fig:gallery_abc}. To further evaluate our model, we perform stress tests by meshing novel shapes from the Thingi10k dataset~\cite{zhou2016thingi10k}, with qualitative results shown in Figure~\ref{fig:abc_generalization}. The results demonstrate that our model accurately reconstructs 3D manifold meshes from input point clouds, showcasing its ability to generalize to unseen shapes during training.

\section{Downstream Task: Mesh Repair}
\label{sec:suppl_mesh_repair}
We provide details on the algorithm to repair a partial mesh using our method. We first obtain the geometry context by sampling the point cloud on the whole surface and feeding it into the point cloud encoder. To generate vertex positions on the target region, while maintaining the original vertices on the untouched region, we re-design the vertex sampling process in the vertex diffusion model, following standard diffusion in-painting approaches.

Specifically, after removing the region to be repaired from the original mesh, we denote the partial mesh that we want to complete as $\Mesh^{\text{known}} = (\Vertices^{\text{known}}, \Edges^{\text{known}}, \Faces^{\text{known}})$. To begin the denoising process, we sample \emph{all} vertices from a Gaussian distribution for time step $T$: $\Vertices_T\sim \mathcal{N}(\mathbf{0}, \mathbf{I})$. At each denoising step $t$, we first align $\Vertices_t$ with the known vertices $\Vertices^{\text{known}}$ to obtain a mask $\mathbf{m}$ which takes value $1$ if the vertex is in $\Vertices_t$ and $0$ otherwise --- see paragraph below. We feed $\Vertices_t$ into our point cloud diffusion model: 
$\Vertices_{t-1}^{\text{unknown}} \sim \mathcal{N}\left(\mu_{\theta}(\Vertices_t, t), \Sigma_{\theta}(\Vertices_t, t)\right)$, where $\mu_{\theta}, \Sigma_{\theta}$ is the mean and variance prediction from our model, respectively. 
The denoised vertices will be $\Vertices_{t-1} =  \mathbf{m} \odot \Vertices_{t-1}^{\text{known}} + (1 - \mathbf{m}) \odot \Vertices_{t-1}^{\text{unknown}}$, where $\Vertices_{t-1}^{\text{known}} \sim \mathcal{N}\left(\sqrt{\bar{\alpha}_t} \Vertices^{\text{known}}, \left(1 - \bar{\alpha}_t\right)\mathbf{I}\right)$. After sufficiently many denoising steps we have the final vertices $\Vertices_{0}$ that match the original mesh except in the region that has been repaired.

The new connectivity is predicted using our connectivity generation model. In this case, to preserve the existing edges and faces from the partial mesh, we only altered the connectivity between the predicted vertices and the boundary vertices from $\Vertices^{\text{known}}$.

\paragraph{Aligning $\Vertices^{\text{known}}$ and $\Vertices_{t}$} Since it is challenging to align two point sets with different numbers of points for each set, we first append surface points sampled from the masked regions, denoted as $p^{\text{masked}}$, to the known vertices $\Vertices^{\text{known}}$, such that $|\Vertices_t|=|\Vertices^{\text{known}}|$. We then solve the correspondence by first computing a cost matrix $\mathbf{C}$, where each item in the matrix $\mathbf{C}_{ij} = \| \Vertices_{t,j} - \Vertices^{\text{known}}_i \|^2_2$. To determine the one-to-one correspondence, we adopt the same strategy that we use for determining the local ordering (Section 3.3 in the main paper). In short, we apply Sinkhorn normalization \cite{sinkhorn1964relationship} to recover a doubly-stochastic matrix, $\hat{\mathbf{C}}$, representing a \emph{softened} permutation matrix~\cite{adams2011ranking}.  The correspondence can be recovered by computing the optimal unconstrained lowest-cost matching \cite{jonker1988shortest}. 

\clearpage

\end{document}